\documentclass[twoside,leqno,twocolumn]{article}

\newtheorem{definition}{Definition}
\usepackage{amsmath}

\newcommand{\method}{{MEXA-CTP}}
\usepackage{hyphenat}
\usepackage[hidelinks]{hyperref}

\usepackage[letterpaper]{geometry}

\usepackage{ltexpprt}
\usepackage{hyperref}
\usepackage{cite}
\usepackage{amsmath,amssymb,amsfonts}
\usepackage{algorithmic}
\usepackage{graphicx}
\usepackage{textcomp}
\usepackage{xcolor}
\usepackage{amsmath,bm}
\usepackage{pbox}
\usepackage{amssymb}
\usepackage{makecell}
\usepackage{lipsum}
\usepackage{caption, subcaption}
\usepackage{textcomp}
\usepackage{upgreek}
\usepackage{algorithm}
\usepackage{url}
\usepackage{multirow}
\usepackage{enumitem,soul}
\usepackage{mathtools,lipsum}
\usepackage{cuted}
\usepackage{xspace}
\usepackage{booktabs}
\usepackage{titling}
\usepackage{cleveref}
\usepackage{fancyhdr}

\makeatletter
\DeclareRobustCommand\onedot{\futurelet\@let@token\@onedot}
\def\@onedot{\ifx\@let@token.\else.\null\fi\xspace}

\makeatother

\def\BibTeX{{\rm B\kern-.05em{\sc i\kern-.025em b}\kern-.08em
    T\kern-.1667em\lower.7ex\hbox{E}\kern-.125emX}}
\begin{document}

%


\title{\method: Mode Experts Cross-Attention for\\ Clinical Trial Outcome Prediction
}


\author{Yiqing Zhang \hspace{1cm} Xiaozhong Liu \hspace{1cm} Fabricio Murai \\
Worcester Polytechnic Institute\\
100 Institute Rd, Worcester, MA, USA\\
{\tt\small \{yzhang37, xliu14, fmurai\}@wpi.edu}
}

\date{}

\maketitle






\begin{abstract} \small\baselineskip=9pt Clinical trials are the gold standard for assessing the effectiveness and safety of drugs for treating diseases. Given the vast design space of drug molecules, elevated financial cost, and multi-year timeline of these trials, research on clinical trial outcome prediction has gained immense traction. Accurate predictions must leverage data of diverse modes such as drug molecules, target diseases, and eligibility criteria to infer successes and failures. Previous Deep Learning approaches for this task, such as HINT, often require wet lab data from synthesized molecules and/or rely on prior knowledge to encode interactions as part of the model architecture. To address these limitations, we propose a light-weight attention-based model, \method{}, to integrate readily-available multi-modal data and generate effective representations via specialized modules dubbed ``mode experts'', while avoiding human biases in model design. We optimize \method{} with the Cauchy loss to capture relevant interactions across modes. Our experiments on the Trial Outcome Prediction (\textsc{TOP}) benchmark demonstrate that \method{} improves upon existing approaches by, respectively, up to 
11.3\% in F1 score, 12.2\% in PR-AUC, and 2.5\% in ROC-AUC, compared to HINT. Ablation studies are provided to quantify the effectiveness of each component in our proposed method. \textbf{Code can be downloaded from} {\sloppy\url{github.com/murai-lab/MEXA-CTP}.}\end{abstract}


\section{Introduction}\label{sec:intro}
\begin{figure}[t] 
\centering 
\includegraphics[width=\linewidth]{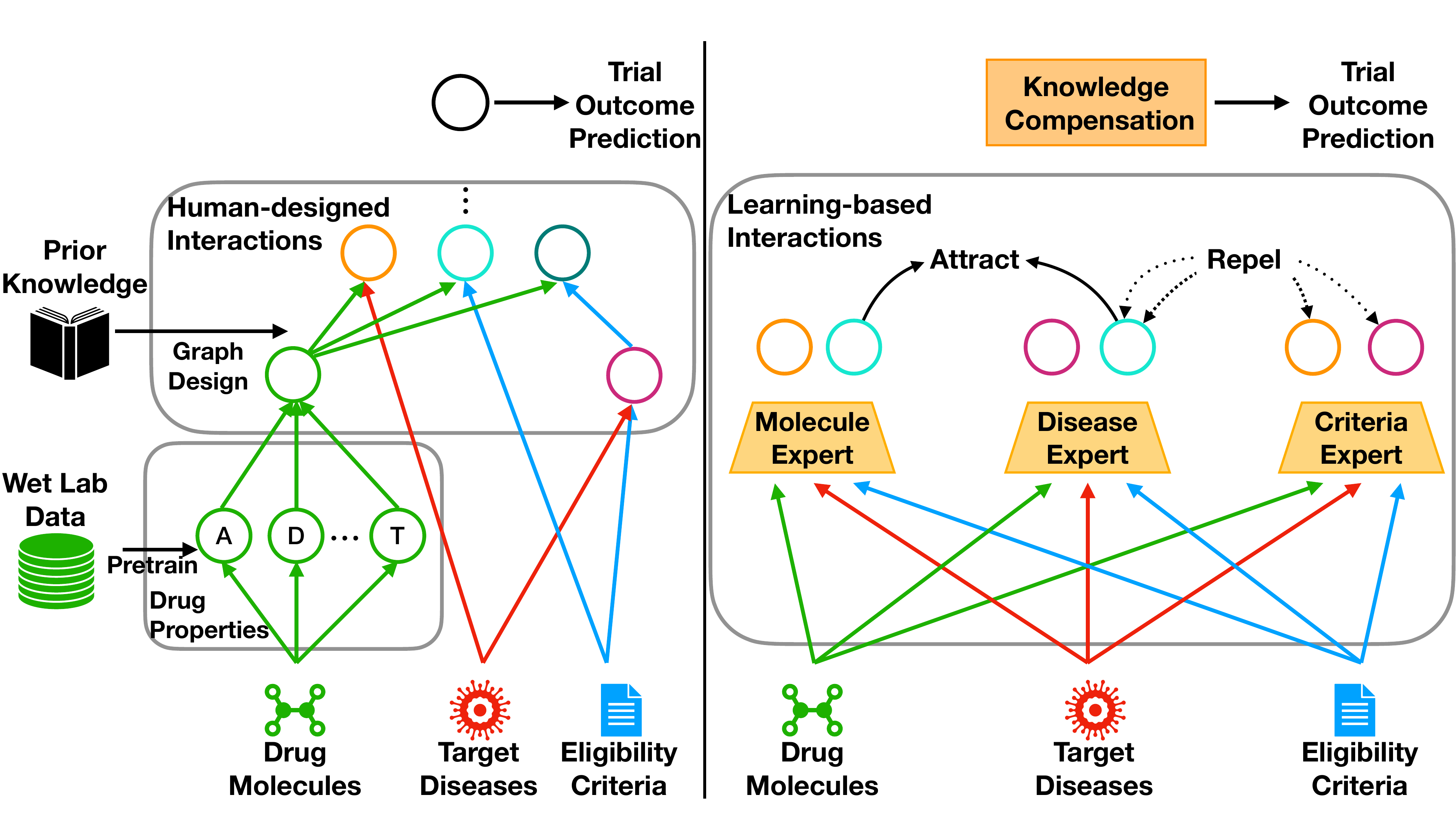}
\caption{(\textit{Left}) Recent work~\cite{fu2022hint} requires wet lab data acquisition for pre-training drug encoders; uses human-designed connections to capture cross-domain interactions. 
(\textit{Right}) \method{} generates domain-specific rich embeddings, filters out irrelevant information, and extracts cross-domain interactions via mode experts. 
} 
\vspace{-6mm}
\label{fig:idea}
\end{figure}

Despite the persistent challenges in clinical trial success, the increasing availability of historical data on clinical trials and extensive knowledge about both approved and failed drugs presents a unique opportunity. Leveraging machine learning/deep learning during the design stages of trials could significantly enhance our ability to predict their likelihood of success. This potential comes at a crucial time, as clinical trials, essential for new drug development, face significant challenges including high costs~\cite{grand2021clinical}, lengthy timelines~\cite{martin2017much}, and a low probability of success, often due to the difficulty of meeting the desired criteria~\cite{ledford2011translational,peace2010clinical}. 
Accurate clinical trial outcome prediction has the potential to shift resources towards trials with a greater chance of positive outcomes, significantly optimizing the drug development landscape. 

Numerous prior efforts have been made to predict clinical trial outcomes and improve trial results, including using 
EEG measurements to identify biomarkers of efficacy, monitor treatment effects in real-time~\cite{chekroud2021promise}, predicting drug toxicity from molecular properties~\cite{hong2020predicting}, and leveraging phase II results to forecast phase III outcomes~\cite{qi2019predicting}. 
In the past years, there has been heightened focus on the broader objective of creating a universal method for predicting trial results across various diseases. 
A very recent method titled Hierarchical INteraction Network (HINT)~\cite{fu2022hint} takes this one step further by incorporating a wide range of multimodal data, including drug molecules, disease information, trial protocols (i.e., inclusion/exclusion criteria) and wet lab data (pharmacokinetics properties of drugs: absorption, distribution, metabolism, excretion, and toxicity), combined through graph neural networks to synthesize and predict outcomes. 
However, these pioneering models face limitations that hinder their broader applicability in accurately forecasting trial results.

\noindent\textbf{Limitations of State-of-the-Art Approaches.} Some prior works aiming to design models for clinical trial outcome prediction rely on biomedical knowledge graphs (BKGs)~\cite{Chen2022} representing the relationship between various biomedical entities. However, it is challenging to incorporate BKGs into clinical trial outcome predictions since public repositories cannot keep up with new discoveries in the literature~\cite{jin2020predicting}, contain uncertain relations and have almost no information on rare diseases. This has severely limited the applications of BKGs to specific diseases, such as COVID-19~\cite{Hsieh2021}. 
In contrast, more general models do not make use of BKGs. Among those, HINT~\cite{fu2022hint} (Fig.~\ref{fig:idea} (\textit{Left})) achieves state-of-the-art performance on the Trial Outcome Prediction (TOP) benchmark. Nevertheless, it has three key limitations: 
\begin{itemize}[nosep, leftmargin=*]
\item \underline{\it Paragraph-level Embedding Limitations.} HINT utilizes paragraph-level embeddings to encode trial protocols. However, this approach does not distinguish between inclusion and exclusion criteria (which respectively specify characteristics that subject \underline{must} and \underline{cannot} have to be eligible to participate), potentially confusing the model about the trial's requirements. Additionally, HINT relies on the combination of  ClinicalBERT~\cite{huang2019clinicalbert} and sliding window~\cite{he2021swacg}, thus overlooking subtle nuances and interconnections between sentences within a paragraph, potentially missing critical information.

\item \underline{\it Reliance on hard-to-acquire data.} Drug molecules are represented through their estimated pharmacokinetics properties (absorption, distribution, metabolism, excretion and toxicity) using a pre-trained model~\cite{ferreira2019admet}. However, this approach is hindered by the need for extensive wet lab data to label these pharmacokinetics properties, which is expensive to acquire and unavailable for molecules that have not yet been synthesized. Additionally, when new drugs becomes available over time, the pretrained model requires retraining to incorporate this new knowledge. 

\item \underline{\it Human biases in neural network design.} Designing neural networks by connecting their modules based on human intuition regarding what they represent and how they should interact is a widely common approach \cite{fu2022hint, harrer2019artificial, yin2012clinical}. Yet, this practice can limit the network's capacity to accurately represent complex relationships within data. Furthermore, it can potentially enforce human biases in the model, ultimately undermining its performance.
\end{itemize}

\noindent\textbf{Our Approach.} 
We introduce a pioneering approach titled \underline{M}ode \underline{Ex}perts Cross-\underline{A}ttention for \underline{C}linical \underline{T}rial Outcome \underline{P}rediction (\method{}) -- illustrated in Fig.~\ref{fig:idea} (\textit{Right}). \method{} leverages the Cauchy loss~\cite{mlotshwa2022cauchy} during optimization, so that the resulting model can use masked cross-attention to selectively combine rich representations extracted via modality-specific modules (referred as \textbf{mode experts}), thus incorporating data representing drug molecules, disease information, and trial protocols in a holistic way. Additionally, \method{} employs a normalized temperature-scaled cross-entropy (NT-Xent)\cite{sohn2016improved} loss to further refine the knowledge captured in the learned representations, culminating in a robust model capable of leveraging cross-domain information.
The proposed method outperforms the current state-of-the-art results on the Trial Outcome Prediction (\textsc{TOP}) benchmark while circumventing the dependence on hard-to-use or expensive-to-acquire data such as BKGs and wet lab data.
\noindent\textbf{Our main contributions are:} 
\begin{itemize}[nosep, leftmargin=*]
\item We introduce \method{}, a new method that integrates multi-modal data based on the concept of mode experts, and is optimized with Cauchy and contrastive losses to capture the relevant drug-disease, drug-protocol, and disease-protocol interactions,  without resorting to hand-crafted structures.

\item We evaluate \method{} against several baselines including the SOTA method, HINT~\cite{fu2022hint}, using real world data from phase I, II and III clinical trials. \method{} yields up to 11.3\%, 12.2\% and 2.5\% of improvement respectively in terms of F1, PR-AUC, and ROC-AUC compared with HINT.

\item We conduct a case study and an ablation study to demonstrate the contribution of key components of \method{} to its prediction power. 
\end{itemize} 

\section{Definitions and Notation}\label{sec:overview}
Before introducing \method{}, we formally define the key components of a (drug-based) clinical trial along with the notation used in this paper.

\begin{definition} \textbf{Drug Molecule} refers to a specific category of pharmaceutical compounds and chemical substances designed to produce pharmacological effects on a target disease.We denote drug molecules without considering the quantities or percentuals\footnote{In clinical trial data, ingredients' quantities or percentuals are not always informed~\cite{umscheid2011key}, as the main focus is on the presence of molecules relevant to the study.} by \begin{equation}
    \mathcal{M}^{(j)} := \{m_1^{(j)}, m_2^{(j)}, \cdots, m_{M_j}^{(j)}\}; \quad j \in [1..N],
\end{equation}
where $m_i^{(j)}; i \in [1..M_j]$ is an active pharmaceutical ingredient (encoded using SMILES representation) of the pharmaceutical compound in clinical trial $j$.
\end{definition}

\begin{definition} \textbf{Target Diseases} are coded using the ICD-10 (International Classification of Diseases, 10$^\mathrm{\textit{th}}$ revision) system, which provides a hierarchical structure for categorizing diseases based on their characteristics. We denote the set of target diseases as
\begin{equation}
    \mathcal{D}^{(j)} := \{d_1^{(j)}, d_2^{(j)}, \cdots, d_{D_j}^{(j)}\}; \quad j \in [1..N],
\end{equation}
where $d_i^{(j)}; i \in [1..D_j]$ is an ICD-10 diagnosis code in clinical trial $j$.
\end{definition}

\begin{definition} \textbf{Eligibility Criteria}, subdivided in inclusion and exclusion criteria, are the specific requirements that individuals must meet to participate in a clinical trial (e.g., age, gender, presence/absence of certain medical conditions, etc). Both are usually specified as a bulleted list of textual statements. The former contains those that must be simultaneously satisfied (AND operator), whereas the latter contains all those that a participator cannot have (OR operator). These criteria are respectively denoted by
\begin{equation}
    \begin{split}
        \mathrm{IC}^{(j)} &:= \{ \mathrm{ic}_1^{(j)}, \mathrm{ic}_2^{(j)}, \cdots, \mathrm{ic}_{{\mathrm{IC}}_j}^{(j)}\}, \; \\
        \mathrm{EC}^{(j)} 
        := &\{ \mathrm{ec}_1^{(j)}, \mathrm{ec}_2^{(j)}, \cdots, \mathrm{ec}_{{\mathrm{EC}}_j}^{(j)}\}; j \in [1..N],
    \end{split}
\end{equation}
where $\mathrm{IC}^{(j)}$, $\mathrm{EC}^{(j)}$ are, respectively, the statement-level inclusion and exclusion criteria in clinical trial $j$, with cardinalities $\mathrm{IC}_j$ and $\mathrm{EC}_j$.
\end{definition}

\begin{definition} \textbf{Clinical Trial} is the process of testing one or more drug molecules on a group of eligible participants to determine its safety or its efficacy in treating a disease. The information regarding a clinical trial $j$ prior to its roll-out is denoted by
\begin{equation}
    \mathcal{X}^{(j)} := \langle\mathcal{M}^{(j)}, \mathcal{D}^{(j)}, \mathcal{C}^{(j)}\rangle; \quad j \in [1..N],
\end{equation}
where $N$ is the total number of clinical trials in the dataset.
\end{definition}
\begin{definition} \textbf{(ML-based) Clinical Trial Outcome Prediction} consists of learning a model $f$ parameterized by $\theta$ which outputs a prediction
\begin{equation}
    \hat{y}^{(j)} = f(\mathcal{X}^{(j)}, \theta);\quad j \in [1..N],
\end{equation}
\end{definition}
for the true outcome of trial $j$, which is typically a binary label $y^{(j)}$. A positive outcome $y^{(j)}=1$ indicates that the drug was effective or safe, and a negative outcome $y^{(j)}=0$ indicates otherwise. 

In order to attain good performance, the model must learn good representations for drug molecules, target diseases, and eligibility criteria, and capture complex relationships among these factors.

\section{Proposed Method}\label{sec:method}
\begin{figure*}[t!] 
\centering 
\includegraphics[width=0.9\linewidth]{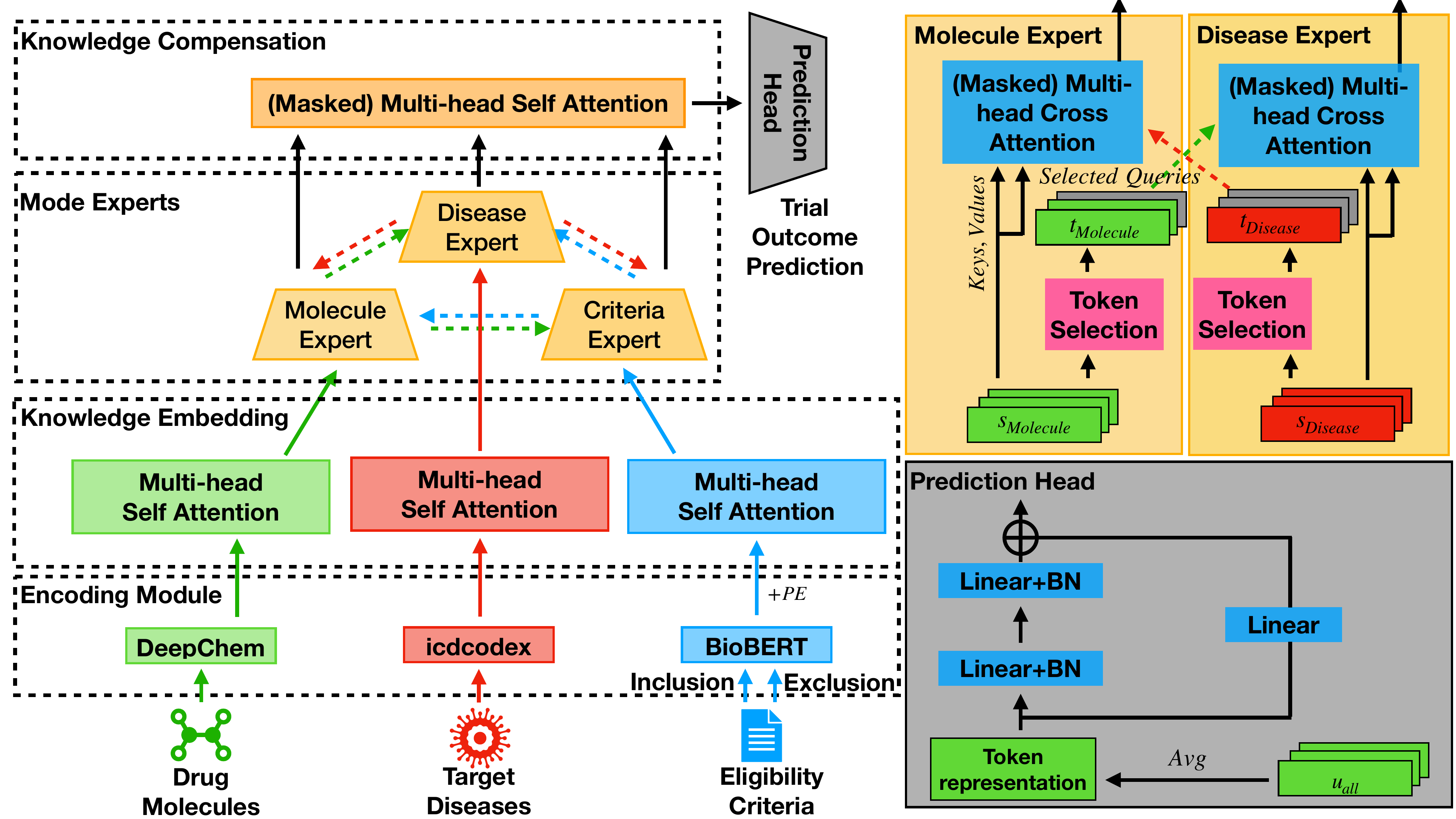}
\caption{(\textit{Left}) \method{} model comprises four stages (from bottom to top): (i) Encoding Modules generate initial embeddings for each mode, (ii) Knowledge Embedding Models further enrich the information, (iii) Mode Experts capture relationships between different modes, and (iv) Knowledge Compensation Module enhances the interactions. (\textit{Top Right}) Example of how two mode experts interact to fuse information from molecule and disease domains; $u$, $s$ and $t$ indicate single tokens from the token set. (\textit{Bottom Right}) We use a residual network for final outcome prediction. Further details are provided in Section~\ref{sec:method}.} 
\label{fig:model}
\end{figure*} 

To tackle the limitations of the existing techniques, we introduce \method{}. It consists of the following four stages (see Fig.~\ref{fig:model}). \underline{Stage 1}: The \textbf{encoding module} leverages modern, publicly-available encoders to extract raw representations for data from each mode (i.e., drug, disease, and eligibility criteria). \underline{Stage 2}: The \textbf{knowledge embedding module} learns how to extract deep features from each mode utilizing the multi-head self-attention mechanism. \underline{Stage 3}: The \textbf{mode experts module}  learns how to capture interactions between drug-disease, drug-criteria, and disease-criteria pairs in a self-supervised fashion. \underline{Stage 4}: The \textbf{knowledge compensation module} fuses all the information provided by the mode experts to predict the outcome of the clinical trial. Below we provide additional details.

\subsection{Stage 1: Encoding Module.}
This module processes data from multiple modalities/modes, including drug molecules, disease information and eligibility criteria. Specifically, we utilize DeepChem~\cite{ramsundar2018molecular} to transform drug molecules, represented as SMILES strings, into drug molecules embeddings, $\textsc{icdcodex}$\footnote{\url{https://github.com/icd-codex/icd-codex}} to map ICD-10 codes to target diseases embeddings, and BioBERT~\cite{lee2020biobert} to generate separate embeddings for the statement-level inclusion and exclusion criteria. We denote these initial embeddings respectively by $U_{\mathcal{M}^{(j)}}$, $U_{\mathcal{D}^{(j)}}$, $U_{\mathrm{IC}^{(j)}}$ and $U_{\mathrm{EC}^{(j)}}$. Details can be found in \cref{appd: encoding}. 
This approach enables us to capture the diverse aspects of the input data and extract meaningful representations for downstream tasks in clinical trial outcome prediction.

\subsection{Stage 2: Knowledge Embedding Module.}
This module enriches the information within each mode, contributing to more meaningful representations. By leveraging mode-specific knowledge and inter-relationships, this module enhances the overall understanding of drug molecules, disease information, and eligibility criteria. For drug molecules and target diseases, we utilize multi-layer transformer encoders to get enriched drug molecules embeddings  $U^{+}_{\mathcal{M}^{(j)}}$ and target diseases embeddings $U^{+}_{\mathcal{D}^{(j)}}$ separately. However, for inclusion and
exclusion criteria, they originate from
highly structured text inputs. By accounting for the statement order, the model can better prioritize and interpret the information in the text.
To capture the statement order as a key feature, we enhance the model by adding sinusoidal positional embeddings to the statements' embeddings. 
We utilize a multi-layer transformer encoder with a siamese design~\cite{chicco2021siamese} to capture information from both inclusion and exclusion criteria to get $U^{+}_{\mathrm{IC}^{(j)}}$ and $U^{+}_{\mathrm{EC}^{(j)}}$. Last, we concatenate the outputs of the inclusion and exclusion criteria to get the enriched embedding of eligibility criteria $U^{+}_{\mathcal{C}^{(j)}} = \textsc{Concatenate}(U^{+}_{\mathrm{IC}^{(j)}}, U^{+}_{\mathrm{EC}^{(j)}})$. Details can be found in~\cref{appd: embedding}.

\subsection{Stage 3: Mode Experts Module}. We model interactions between drug-disease, drug-criteria, and disease-criteria pairs using three mode experts. Each mode expert performs two main functions: (i) the token selection function chooses essential output tokens to serve as queries for constructing interactions with other mode experts, and (ii) the cross-attention function generates values and keys from \textbf{source tokens} $S$ to respond to queries based on \textbf{target tokens} $T$ from other mode experts, thereby accounting for interactions.

\subsubsection{Token Selection.} The token selection for these interactions is based on filtering noisy information, which helps the model attend to the most relevant tokens for predicting each trial's outcome. We utilize a projection layer to predict the probability of each token being selected by the mode expert:
\begin{equation}
    p(S) = \mathrm{sigmoid} (SW_p),
\end{equation}
where $S \in \{ U^{+}_{\mathcal{M}^{(j)}}, \,U^{+}_{\mathcal{D}^{(j)}}, \,U^{+}_{\mathcal{C}^{(j)}} \}$. Intuitively, $p_{S}$ represents the confidence that token $S$ should be used for querying other experts. Note that all tokens $S$ are used for generating keys and values for cross-mode attention.

We employ two mechanisms to control the quality of the output tokens $T$, to be sent to the cross-attention layer.
First, we apply a hard margin to mask tokens with confidence levels lower than a specified threshold $t$; then we apply a soft margin which modulates $S$ with its confidence level $p_S'$:
\begin{equation}
    p'(S) = I_{p(S) \leq t} \odot p(S), \quad     T = p'(S) \odot S,
\label{equ:token_selection_1_and_2}
\end{equation}
where $I_{p(S) \leq t}$ is an indicator function, operator $\odot$ denotes element-wise multiplication. The soft margin step allows for a more gradual and continuous masking of tokens based on their confidence levels, contributing to the overall learning process.

To ensure that the probability distribution focuses only on the minimum number of necessary tokens and prevent the model from overfitting to noisy or less informative features, we introduce the Cauchy loss~\cite{mlotshwa2022cauchy},
\begin{equation}
    L_{\mathrm{cauchy}} = \sum_{s \in S} \log\left(1 + \frac{p_s^2}{\epsilon}\right),
\end{equation}
where $\epsilon$ is a hyperparameter to control the confidence level of each token. This loss can be jointly optimized with other loss functions 
to control the quality of the output tokens.

Each of the three mode experts shown in Fig.~\ref{fig:model} (\textit{Left}) are responsible for selecting queries to be forwarded to the other mode experts. These queries are denoted by $T_{\mathcal{M}^{(j)}}$, $T_{\mathcal{D}^{(j)}}$, and $T_{\mathcal{C}^{(j)}}$. Fig.~\ref{fig:model} (\textit{Top Right}) illustrates the interactions between molecule and disease experts in detail.

\subsubsection{Cross-Attention.} 
The cross attention aims to build interactions between two modes.
A mode expert from mode $D \in \{\mathcal{M}, \mathcal{D}, \mathcal{C}\}$ combines information from its own mode (source tokens $S_D$) with information from another mode $D' \neq D$ received from the respective expert (target tokens $T_{D'}$) via cross-attention. More precisely, this operation is expressed as 
\begin{equation}
\overrightarrow{I}_{{D'D}} = \mathrm{softmax}\left(\frac{T_{D'}W_D^Q (S_{D}W_D^K)^\top}{\sqrt{d_k}}\right) S_{D}W_D^V,
\end{equation}
where the arrow direction signifies \textit{$D'$ forwards target tokens to $D$}. This mechanism is repeated for all mode pairs, allowing the model to capture interactions between different modes and learn meaningful representations. For conciseness, we adopt shorter notations for those pairs: $I_{md},\, I_{dm},\, I_{cd},\, I_{dc},\, I_{mc},\, I_{cm}$.

\subsubsection{Self-supervised Learning.}
We leverage the output tokens from mode experts with self-supervised learning to build better semantic-level representations. Unlike general contrastive learning methods in large language models (LLMs), we define anchors at the semantic level. For instance, we obtain pairs representing interactions between molecules and diseases from $I_{md}$, and the reciprocal pairs from $I_{dm}$. We treat these pairs as positive samples and consider other pairs as negative samples by defining our contrastive loss as follows:
\begin{equation}
    \begin{split}
    L_{\mathrm{contrastive}} &= \\ -\sum_{\mathrm{pair_+} \in \mathcal{P}_+} &\log \left( \frac{ \mathrm{exp}(\mathrm{sim}(\mathrm{pair}_+)/\tau)} {\sum_{\mathrm{pair_{all}} \in \mathcal{P}_\mathrm{all}} \mathrm{exp}(\mathrm{sim}(\mathrm{pair_{all}})/\tau)} \right),
    \end{split}
\label{equ:contrastive}
\end{equation}
where $\mathcal{P}_+ = \{(I_{md}, I_{dm}), (I_{cm}, I_{mc}), (I_{cd}, I_{dc}) \}$,  $\mathcal{P}_\textrm{all} = \{(I_{D'D},I_{\delta'\delta} ) | \, (D'\,, D\,, \delta'\,,\delta) \in \{m\,,d\,,c\}^4\, \land (D' \neq D) \land (\delta' \neq \delta) \land (D' \neq \delta' \lor D \neq \delta) \}$, $\mathrm{sim}(.)$ is the cosine similarity function, and $\tau$ is the temperature of the contrastive loss.

\subsection{Stage 4: Knowledge Compensation Module.} The aim of the knowledge compensation module is to enhance interactions between molecules, diseases, and criteria. 
Following a self-attention encoder, we concatenate the output tokens from each expert and average them across the sequence dimension, and forward them through a projection head with a sigmoid function to predict the probability that the trial is successful:
\begin{equation}
    \hat{y} = \sigma(f_{\mathrm{pred}}(\textsc{Average}(I_{\mathrm{all}}))),
\end{equation}
where $I_{\mathrm{all}} = \textsc{Concatenate}(I_{md},\, I_{dm},\, I_{cd},\, I_{dc},\, I_{mc},\, I_{cm})$.
A class-weighted binary cross-entropy (wBCE) loss is used for guiding the model training due to the imbalance of negative/positive pairs:
\begin{equation}
    L_\mathrm{cls} = -\omega_{0} y\log\hat{y} - \omega_{1} (1-y)\log(1-\hat{y}),
\end{equation}
where $\omega_{0}$ ($\omega_{1}$) is the fraction of negative (positive) labels in the training set.

\subsection{Loss function}  The final loss is a combination of classification, Cauchy and contrastive losses:
\begin{equation}
    L = L_{\mathrm{cls}} + \lambda_1 L_{\mathrm{cauchy}} + \lambda_2 L_{\mathrm{contrastive}},
\label{equ:loss}
\end{equation}
where $\lambda_1$ and $\lambda_2$ are hyperparameters which help balance the Cauchy loss and contrastive loss. We conduct a grid search to tune $\lambda_1$ and $\lambda_2$ on the validation set.

\section{Experiments}\label{sec:experiment}
We carefully follow the same experimental settings as in the Trial Outcome Prediction (\textsc{TOP}) benchmark~\cite{fu2022hint} to evaluate our model. In addition, to demonstrate the way the mode expert works and gain intuition on its importance, we visualize the token selection ratios. Last, to assess the capabilities of the learning components, we conduct ablation studies on different parts of our model.

\subsection{Experimental Settings.}

\textit{Dataset.} To the best of our knowledge, the \textsc{TOP} benchmark includes all available trials and their outcomes from DrugBank\footnote{\url{https://www.drugbank.com/use_cases/ml-drug-discovery-repurposing}}. To evalute the proposed model, we use the \textsc{TOP} benchmark and follow the same strategy for splitting the train, validation, and test sets. Specifically, for each phase, we split the dataset based on the start day of the clinical trials, allocating earlier trials to the train set and using later trials for model evaluation. 
During training, we randomly select $15\%$ of the training samples as the validation set to monitor the model's performance and tune hyperparameters. This protocol is part of the TOP's experimental design, ensuring consistency and fair comparisons with other studies. For statistics of the data splits, please refer to Table~\ref{statistics}.

\begin{table*}[t!]
\caption{\textit{TOP} benchmark dataset statistics.} 
\label{statistics}
\centering
\setlength\tabcolsep{15pt}
\begin{tabular}{ccccl}
\toprule
Phase &  Training & Test & Success Ratio & Split Day\\
\midrule
I & 1,088 & 312 & 70\% & Aug 13, 2014\\
II & 2,611 & 789 & 33\% & March 20, 2014\\
III & 4,313 & 1,147 & 30\% &  April 7, 2014\\
\bottomrule
\end{tabular}
\end{table*}

\noindent \textit{Our Model.} We stack 2 attention layers in each attention block. For each attention layer, we utilize 2 attention heads, with an embedding size of 16 and explosion size of 32. We optimize our model via Adam optimizer using default $\beta_1$ and $\beta_2$. We set the mini-batch size to 64 and use a fixed learning rate of $5e^{-2}$ for training.

\noindent \textit{Baselines.} We compare \method{} with several baselines, including Logistic Regression (\textbf{LR}), Random Forest (\textbf{RF}), k-Nearest Neighbor + Random Forest (\textbf{kNN+RF}), \textbf{XGBoost}, Adaptive Boosting (\textbf{AdaBoost}), a 3-layer Feed-Forward Neural Network (\textbf{FFNN}) and \textbf{HINT}~\cite{fu2022hint}. For the eligibility criteria input, we utilize paragraph-level embeddings, while other settings were adapted directly from~\cite{fu2022hint}. 

\noindent \textit{Training \& Testing Protocols.} We follow the same training and testing settings for the TOP dataset as in previous works. For each phase, after determining the best hyperparameters using the validation loss, we re-train our model on the training and validation sets combined. For test, we use bootstrapping to evaluate the model performance 10 times on a random selection of 80\% of the test data to report the mean and standard deviation of the evaluation metrics. 

\subsection{Experiment Results.}\label{sec:results}

Table~\ref{results}shows the comparison results for each phase, including standard deviation values obtained via bootstrapping. For all phases, \method{} consistently achieves the best performance in terms of F1, PR-AUC and ROC-AUC, immediately followed by HINT.

The performance gaps over HINT are larger in terms of the first two metrics: between 5.3 and 19.2\% for F1; 4.1 and 27.9\% for PR-AUC; and 1.1 and 3.5\% for ROC-AUC. For comparison purposes, we compute simple (unweighted) averages of their performance differences across phases. On average, \method{} achieves 11.3\%, 12.2\% and 2.5\% higher F1, PR-AUC and ROC-AUC than HINT.

\begin{table*}[t!]
\caption{Experimental results for outcome prediction for phase I, II and III trials. Results correspond to averages and standard deviations over 10 bootstrap samples.}
\centering
\setlength\tabcolsep{15pt}
\begin{tabular}{clccc}
\toprule
Phase & Method &  F1 & PR-AUC & ROC-AUC\\
\midrule
\multirow{8}{*}{I}&LR & .495$\pm$.011 & .513$\pm$.015 & .485$\pm$.015\\
&RF & .499$\pm$.016 & .514$\pm$.015 & .542$\pm$.016\\
&KNN+RF & .621$\pm$.018 & .513$\pm$.004 & .528$\pm$.009\\
&XGBoost & .624$\pm$.016 & .594$\pm$.015 & .539$\pm$.012\\
&AdaBoost & .633$\pm$.015 &.544$\pm$.010 & .540$\pm$.012\\
&FFNN & .634$\pm$.027 & .576$\pm$.020 &.550$\pm$.020\\
&HINT & .598$\pm$.011& .581$\pm$.021&.573$\pm$.024\\
&\method{} & \textbf{.713}$\pm$.027 & \textbf{.605}$\pm$.014 & \textbf{.593}$\pm$.012\\
\hline
\multirow{8}{*}{II}&LR & .527$\pm$.016 & .560$\pm$.012 & .559$\pm$.006\\
&RF & .463$\pm$.011 & .553$\pm$.017 & .626$\pm$.009\\
&KNN+RF & .624$\pm$.011 & .573$\pm$.022 & .560$\pm$.017\\
&XGBoost & .552$\pm$.005 & .585$\pm$.015 & .630$\pm$.003\\
&AdaBoost & .583$\pm$.008 &.586$\pm$.011 & .603$\pm$.002\\
&FFNN & .564$\pm$.014 & .589$\pm$.015 &.610$\pm$.012\\
&HINT & .635$\pm$.011& .607$\pm$.012&.621$\pm$.015\\
&\method{} & \textbf{.695}$\pm$.008 & \textbf{.635}$\pm$.015 &\textbf{.638}$\pm$.005\\
\hline
\multirow{8}{*}{III}&LR & .624$\pm$.013 & .553$\pm$.011 & .600$\pm$.028\\
&RF & .675$\pm$.018 & .583$\pm$.024 & .643$\pm$.023\\
&KNN+RF & .670$\pm$.018 & .587$\pm$.016 & .643$\pm$.024\\
&XGBoost & .694$\pm$.017 & .627$\pm$.009 & .668$\pm$.014\\
&AdaBoost & .722$\pm$.014 &.589$\pm$.015 & .624$\pm$.013\\
&FFNN & .625$\pm$.017 & .572$\pm$.020 &.620$\pm$.023\\
&HINT & .814$\pm$.013& .603$\pm$.014&.685$\pm$.023\\
&\method{} & \textbf{.857}$\pm$.007 & \textbf{.771}$\pm$.016 &\textbf{.693}$\pm$.025\\
\bottomrule
\end{tabular}
\label{results}
\end{table*}

\subsection{Token Usage for Token Selection.}
\begin{figure}[t] 
\centering 
\includegraphics[width=.9\linewidth]{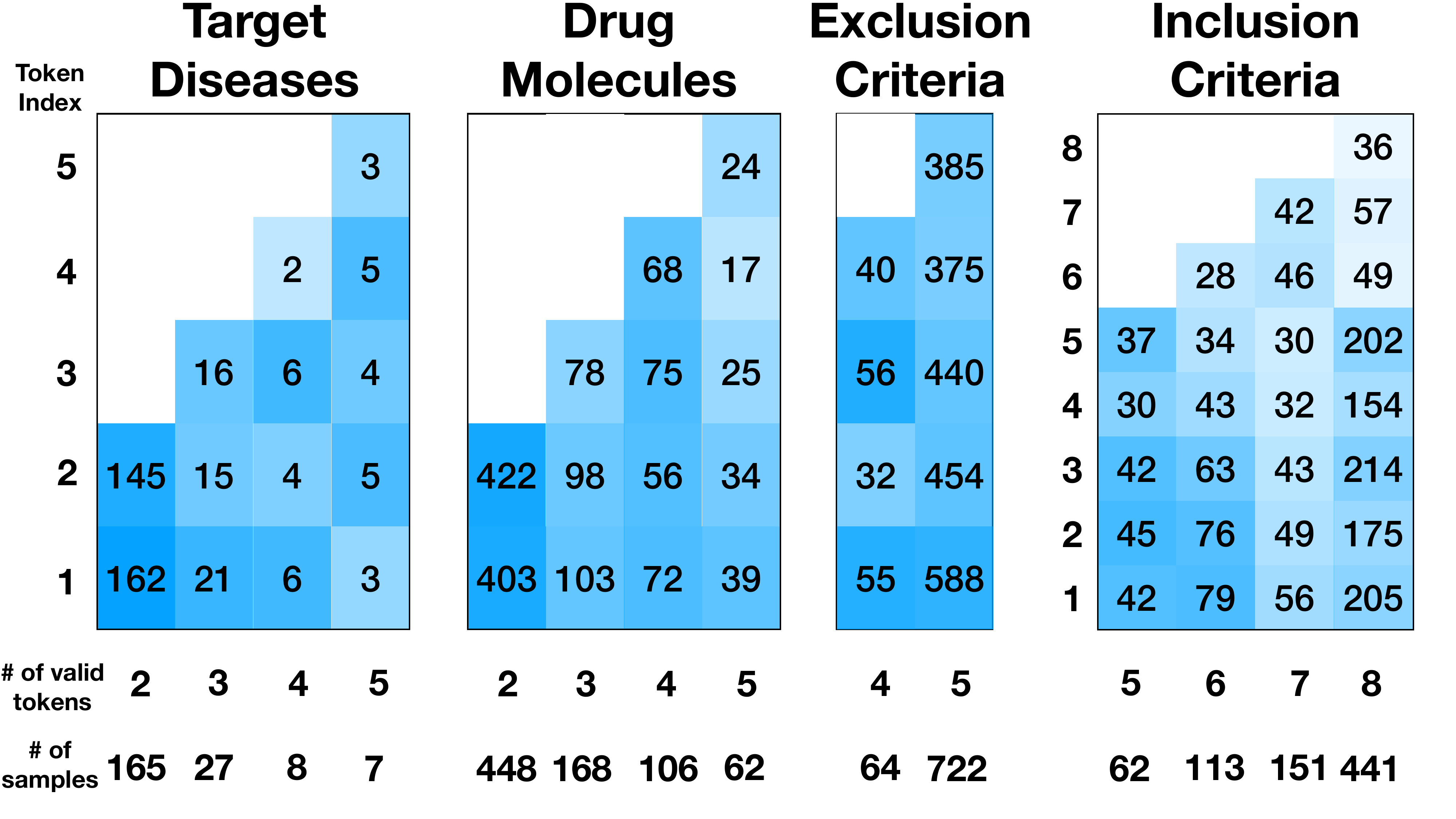}
\caption{Token selection frequency per index, grouped by number of valid tokens $x$. Color indicates ratio between selection frequency and number of samples. For exclusion (resp.\ inclusion) criteria, columns $x < 4$ (resp.\ $x < 5$) excluded due to small number of samples. Columns $x = 1$ omitted since token is always selected.} 
\label{fig:usage}
\end{figure}

Regarding the analysis of the effectiveness of token selection in Section~\ref{sec:method}, we conduct an in-depth examination of that process by visualizing the selected rate of tokens by mode experts.
In Fig.~\ref{fig:usage}, we visualize the token selection rate by different mode experts. While our work includes only one criteria expert to handle both inclusion and exclusion criteria, for clarity, we present the results in separate figures. The disease expert and molecule expert show almost uniform token selection distribution. However, for the criteria expert, tokens with smaller token indexes have a higher chance of being selected, which is consistent with the intuition that criteria's order reflect importance.

\subsection{Ablation Studies.}
We conduct ablation studies for assessing (i) the effectiveness of \method{}'s information aggregation method at the statement level, (ii) the contribution of positional embedding at statement level, (iii) the impact of using the Cauchy loss for selecting meaningful tokens, (iv) the influence of using the contrastive loss for cross-mode representation learning, (v) and token selection methods. In these studies we use the best hyperparameters for our model in phase III obtained in Section~\ref{sec:results}, unless stated otherwise.

\subsubsection{Information Aggregation at the Statement Level.}\label{sec:ablations_ia}
Aiming to obtain the best representation for inclusion/exclusion criteria, we explore three methods for aggregating information at the statement level in Table~\ref{tab:ab}: directly using the [CLS] token, averaging all tokens (except for [CLS] and [SEP] tokens), and summing all tokens (except for [CLS] and [SEP] tokens).
\begin{table*}[t!]
\caption{Ablation studies for statement-level information aggregation, positional embeddings, and loss function variants. Defaults: aggregation using the [CLS] token; with positional embeddings; loss as defined in Eq.~\eqref{equ:loss}. 
}
\centering
\setlength\tabcolsep{5pt}
\begin{tabular}{llccc}
\toprule
Ablations & Method &  F1 & PR-AUC & ROC-AUC \\
\midrule
\multirow{3}{*}{Aggregation}&\method{} (\textit{Avg}) & .776$\pm$.014 & .722$\pm$.014 & .487$\pm$.018\\
&\method{} (\textit{Sum}) & .711$\pm$.015 & .690$\pm$.012 & .445$\pm$.023\\
&\method{} & \textbf{.857}$\pm$.007 & \textbf{.771}$\pm$.016 &\textbf{.693}$\pm$.025\\
\midrule
\multirow{2}{*}{$\mathrm{PE} $}&\method{} (w/o $\mathrm{PE}$) & .846$\pm$.014 & .755$\pm$.011 & .691$\pm$.014\\
&\method{} & \textbf{.857}$\pm$.007 & \textbf{.771}$\pm$.016 &\textbf{.693}$\pm$.025\\
\midrule
\multirow{4}{*}{Loss}&BCE only & .700$\pm$.011 & .744$\pm$.012 & .473$\pm$.015\\
&BCE + Cauchy & .790$\pm$.008 & .753$\pm$.012 & .641$\pm$.018\\
&BCE + contrastive &.810$\pm$.012 & .722$\pm$.014 & .643$\pm$.026\\
&\method{} & \textbf{.857}$\pm$.007& \textbf{.771}$\pm$.016 &\textbf{.693}$\pm$.025\\
\midrule
\multirow{3}{*}{Token Selection}& Random& .328$\pm$.018&.466$\pm$.015&.544$\pm$.017\\
& ALL & .854$\pm$.012&.696$\pm$.014&\textbf{.695}$\pm$.027\\
&\method{} & \textbf{.857}$\pm$.007& \textbf{.771}$\pm$.016 &.693$\pm$.025\\

\bottomrule
\end{tabular}
\label{tab:ab}
\end{table*}

As shown in Table~\ref{tab:ab}, using only the embedding of the first token ([CLS]) to represent the statement-level criteria yields the best performance. In addition to showing significant improvements in comparison to the other aggregation methods, it also leads to slightly smaller computational costs.

\subsubsection{Positional Embedding at Statement level.} To understand whether the order is an important feature of eligibility criteria, we conduct ablation study by experimenting with and without positional embedding for inclusion and exclusion criteria in Table~\ref{tab:ab}. \method{} with positional embeddings shows improvements on F1, PR-AUC and ROC-AUC.

\noindent \textit{Cross-mode Representation Learning.} We introduce the Cauchy loss for token selection and the contrastive loss for cross-mode representation learning. We use $\lambda_1$ and $\lambda_2$ to control the magnitude of the loss, respectively.  To examine the gains originating from each of these losses and compare them with their combined usage in \method{}, we evaluate three variants: wBCE only ($\lambda_1=0$ and $\lambda_2=0$), wBCE with Cauchy ($\lambda_1=0.05$ and $\lambda_2=0$), and wBCE with constrastive ($\lambda_1=0$ and $\lambda_2=0.04$). 
In summary, the combination of the three losses in Eq.~\eqref{equ:loss} significantly outperforms all the simplified loss variants.

\noindent \textit{Token Selection.} To demonstrate that our token selection method is selecting both a sufficient quantity and the appropriate tokens, we evaluated our model using two distinct strategies on the complete phase III dataset. First, we compared our model results with its performance using all available tokens. The model's performance with our design is comparable to using all tokens in terms of ROC-AUC. Additionally, \method{} exhibited slightly better performance with using all tokens in both F1 score and PR-AUC, indicating that our token selection approach is effective and does not discard useful tokens. Next, we tested the model with tokens randomly selected to match the number of tokens chosen by our hard margin method. We observed a dramatic drop in performance when using the random strategy, demonstrating that the token selection strongly contributes to \method's performance.

\subsection{Complexity Analysis} 
As a transformer-based model \cite{hu2023self, hu2024only}, \method{}'s time complexity for training is $\mathcal{O}\left( d n^2 \right)$, where $d$ is the hidden size of the attention model (we set $d=32$), and $n$ is the maximum number of tokens considered by the model. Given the maximum number of drug molecules, target diseases, inclusion criteria, and exclusion criteria, $n \leq 2 \cdot (5+5+8+5) = 46$.

\section{Related Work}\label{sec:related}
\noindent \textit{Clinical Trial Matching.} Many studies have focused on learning patient retrieval and enrollment information for predicting individual patient outcomes within clinical trials, rather than making overall predictions about trial success.
Doctor2Vec~\cite{biswal2020doctor2vec} learns representations for medical providers from EHR data, and for trials from their descriptions and categorical information, in order to address data insufficiency issues such as trial recruitment in less populated countries. DeepEnroll~\cite{zhang2020deepenroll} encodes enrollment criteria and patient records into a shared latent space for matching inference. COMPOSE~\cite{gao2020compose} encodes structured patient records into multiple levels based on medical ontology and used the eligibility criteria embedding as queries to enable dynamic patient-trial matching. In contrast, our work focuses on predicting the clinical trial outcome directly based on drug molecules, target diseases, and eligibility criteria.

\noindent \textit{Clinical Trial Outcome Prediction.}
The first works using classic machine learning method for clinical trial outcome prediction focused on one specific trial~\cite{wu2012identifying}.
More recently, there have been numerous efforts to build more general models. Hong et al.~\cite{hong2020predicting} focused on forecasting clinical drug toxicity using features related to drug and target properties, employing an ensemble classifier of weighted least squares support vector regression. RS-RNN~\cite{qi2019predicting} predicted phase III outcomes based on phase II results by considering time-invariant and time-variant variables. EBM-Net~\cite{jin2020predicting} inferred clinical trial outcomes by unstructured sentences from medical literature that implicitly contain PICOs (Population, Intervention, Comparison and outcome). More recently, HINT~\cite{fu2022hint} incorporated drug molecule features, target diseases, and eligibility criteria to build a hierarchical graph (tree). While these studies optimize representation learning for either engineered features or manual-designed graph, our method \method{} predicts clinical trial outcome using same input as HINT but with minimal human efforts.

\section{Conclusion}\label{sec:conclusion}
This paper presents \method{}, a novel lightweight approach for predicting clinical trial outcomes. 
\method{} addresses prior limitations by incorporating statement-level embeddings from eligibility criteria, and integrating multi-modal data via mode experts. By guiding the optimization loss through carefully \-designed loss functions (Cauchy Loss and contrastive loss), our model leverages ``mode expert'' modules to learn interactions across different domains.
Our evaluation on the Trial Outcome Prediction (\textsc{TOP}) benchmark demonstrates that \method{} yields gains of up to 11.3\% in F1, 12.2\% in PR-AUC, and 2.5\% in ROC-AUC metrics compared with the previous SOTA, HINT.



\bibliographystyle{siam}
\bibliography{reference}

\newpage
\appendix
\section{Encoding Module.} \label{appd: encoding}
\subsection{Drug Molecules Embedding.}
We observe that although drug molecules may differ, they often share some SMILES segments. Therefore, we create molecule embeddings by intelligently combining their SMILES segment representations as obtained by DeepChem~\cite{ramsundar2018molecular}. To improve efficiency, we build an embedding dictionary $\textsc{Dict}_\mathrm{emb}$ for each SMILES segment,
\begin{equation}
    \textsc{Dict}_\mathrm{emb} = \{ e_{s_1}, e_{s_2}, \cdots, e_{s_{V}}\}, \quad  e_{s_{k}}= \textsc{DeepChem} (s_k)
\end{equation}
where $s_k; \, k \in [1..V]$ is a SMILES segment and $e_{s_k}$ is the corresponding representation pre-computed by $\textsc{DeepChem}$. Molecules embeddings are represented as a sequence of tokens from $\textsc{Dict}_\mathrm{emb}$, denoted by $U_{\mathcal{M}^{(j)}}$.

\subsection{Target Diseases Embedding.} Each target disease in a trial is represented using the ICD-10 code tree, which expresses how different diseases and related within the disease classification system. We encode the structural information of each code using the $\textsc{icdcodex}$ package:
\begin{equation}
    \begin{split}
    U_{\mathcal{D}^{(j)}} =  
    \{e_{d_1^{(j)}}, e_{d_2^{(j)}}, \cdots, e_{d_{D_j}^{(j)}}\}, &\quad e_{d_{i}^{(j)}} = \textsc{icdcodex} (d_i^{(j)}), \\ i &\in [1..D_j],
    \end{split}
\end{equation}
where $U_{\mathcal{D}^{(j)}}$ represents a sequence of tokens for target disease embeddings.

\subsection{Eligibility Criteria Embedding.} We split the inclusion and exclusion criteria at the statement level based on the keywords ``inclusion'' and ``exclusion''. If these keywords are not found, we consider the entire paragraph as inclusion criteria, and use zero vectors for the exclusion criteria embedding. We use BioBERT~\cite{lee2020biobert} to extract information from each statement, adding [CLS] (classification) and [SEP] (separator) tokens as per the model's convention. The inclusion and exclusion statement-level embeddings are denoted by
\begin{equation}
    \begin{split}
        U_{\mathrm{IC}^{(j)}} = \{ e_{\mathrm{ic}_1^{(j)}}, e_{\mathrm{ic}_2^{(j)}}, \cdots, e_{\mathrm{ic}_{{\mathrm{IC}}_j}^{(j)}}\}, \\
        U_{\mathrm{EC}^{(j)}}
        = \{ e_{\mathrm{ec}_1^{(j)}}, e_{\mathrm{ec}_2^{(j)}}, \cdots, e_{\mathrm{ec}_{{\mathrm{EC}}_j}^{(j)}}\}.
    \end{split}
\end{equation}
We then aggregate the statement-level embeddings within each criteria:
\begin{equation}
        e_{\mathrm{xc}^{(j)}} = \textsc{Agg}(\mathrm{BioBERT}(\mathrm{xc}^{(j))}),
\end{equation}
where $\textsc{Agg}(.)$ is the aggregation method that helps the BioBERT model generate statement-level embeddings, $\mathrm{xc}^{(j)}$ is a statement from inclusion/exlusion criteria, $e_{\mathrm{xc}^{(j)}}$ is its corrsponding embeddings. We explore three different aggregation methods: first token, average, and sum. 

\section{Knowledge Embedding Module.} \label{appd: embedding}
\subsection{Drug Molecules \& Target Diseases.} 
The dimensions of the drug molecules embeddings $U_{\mathcal{M}^{(j)}}$ and target disease embeddings $U_{\mathcal{D}^{(j)}}$ are intentionally kept low to reduce the complexity of the model and improve computational efficiency. However, to enrich the features in each mode and capture more complex relationships, we utilize a multi-layer transformer encoder. This allows us to effectively process and encode the information from the low-dimensional embeddings, enabling the model to learn more nuanced representations that can better capture the characteristics of the drug molecules and target diseases.

Therefore, we get enriched drug molecules embeddings $U^{+}_{\mathcal{M}^{(j)}}$ and target diseases embeddings $U^{+}_{\mathcal{D}^{(j)}}$. We use $d_k=32$ as the dimension size of enriched tokens.

\subsection{Eligibility Criteria.}
We treat inclusion and exclusion criteria separately. 
As they originate from highly structured text inputs,
the order of statements in eligibility criteria can determine the logical flow and requirements for inclusion or exclusion. 
As in this case, the order of statements often reflects the order of importance or relevance in the context of eligibility criteria for clinical trials, with key information typically presented first. Understanding the statement order allows the model to capture these relationships and make more accurate predictions. By considering the statement order, the model can better prioritize and interpret the information in the text.
To capture the statement order as a key feature, we enhance the model by adding sinusoidal positional embeddings to the statements with its corespondent order. 
We utilize a multi-layer transformer encoder with siamese design~\cite{chicco2021siamese} to capture information from both inclusion and exclusion criteria to get $U^{+}_{\mathrm{IC}^{(j)}}$ and $U^{+}_{\mathrm{EC}^{(j)}}$.
Then we merge outputs of the inclusion criteria and exclusion criteria together as the enriched embedding of eligibility criteria $U^{+}_{\mathcal{C}^{(j)}} = \textsc{Concatenate}(U^{+}_{\mathrm{IC}^{(j)}}, U^{+}_{\mathrm{EC}^{(j)}})$.

\section{Baselines.} 
\textbf{ML-based methods.}
\noindent \textit{Implementation.} We utilized scikit-learn packages for all machine learning baselines, including Logistic Regression, Random Forest, k-Nearest Neighbor + Random Forest, XGBoost, Adaptive Boosting, a 3-layer Feed-Forward Neural Network.


\noindent \textit{Data Preprocessing.}  We utilize the same encoding module outlined in \cref{appd: encoding}. For handling missing values, if the method involves k-nearest neighbors, we will generate k clusters using the non-missing values and predict the missing value based on the centroid of the corresponding cluster. For methods that do not incorporate clustering, we will substitute missing values with zeros. Additionally, we will pad and chunk the array to ensure same size, using the normalized array as input for all the baseline models.

\noindent \textit{Optimization.} We applied class weight parameters to recalibrate the loss when the respective functions provided this hyperparameter.  To gain best hyperparameters for each function in each phase, we employed 5-fold cross-validation while training.

\noindent \textbf{HINT.} We strictly follow the HINT paper and their official GitHub\footnote{\url{https://github.com/futianfan/clinical-trial-outcome-prediction}}.

\end{document}